\documentclass[10pt, a4paper]{article}

\usepackage[final]{lrec-coling2024} 
\usepackage{times}
\usepackage{latexsym}
\usepackage{graphicx}
\usepackage{tabularx}
\usepackage[normalem]{ulem}
\usepackage{multirow}
\usepackage{url}
\usepackage[export]{adjustbox} 
\usepackage{ctable} 
\usepackage{float}
 \usepackage{caption}

\title{PerSHOP - A Persian dataset for shopping dialogue systems modeling}

\name{Keyvan Mahmoudi, Heshaam Faili} 
\address{School of Electrical and Computer Engineering, College of Engineering, University of Tehran, Tehran, Iran\\          
         kmahmoudi@ut.ac.ir, hfaili@ut.ac.ir\\
        }

\abstract{
Nowadays, dialogue systems are used in many fields of industry and research. There are successful instances of these systems, such as Apple Siri, Google Assistant, and IBM Watson. Task-oriented dialogue system is a category of these, that are used in specific tasks. They can perform tasks such as booking plane tickets or making restaurant reservations. Shopping is one of the most popular areas on these systems. The bot replaces the human salesperson and interacts with the customers by speaking. To train the models behind the scenes of these systems, annotated data is needed. In this paper, we developed a dataset of dialogues in the Persian language through crowd-sourcing. We annotated these dialogues to train a model. This dataset contains nearly 22k utterances in 15 different domains and 1061 dialogues. This is the largest Persian dataset in this field, which is provided freely so that future researchers can use it. Also, we proposed some baseline models for natural language understanding (NLU) tasks. These models perform two tasks for NLU: intent classification and entity extraction. The F-1 score metric obtained for intent classification is around 91\% and for entity extraction is around 93\%, which can be a baseline for future research.
 \\ \newline  }

\begin{document}

\maketitleabstract

\section{Introduction}
Task-oriented dialogue systems are getting more interesting in research and industry. These systems could work for various tasks like reservation systems, alarm clocks, finding a place on a map, and so on. Shopping dialogue systems are one of the popular areas of these systems where a dialogue system takes a human’s place as a seller and makes conversations with the customers. These systems are supervised, so they need annotated datasets to train. Currently, there are datasets in different fields for task-oriented dialogue systems, such as MultiWOZ \citep{MultiWOZ}, SGD \citep{SGD}, RiSAWOZ\citep{risawoz}, and NLU++ \citep{casanueva-etal-2022-nlu}. In each of these datasets, a suitable amount of data has been collected and annotated. In some cases, it has been presented baselines on these datasets to train and evaluate models in various tasks like natural language understanding, dialogue state tracking, dialogue policy, and natural language generation. Although the datasets mentioned above are suitable for training a model, none of them is in the field of shopping systems. \citep{Parsi} present an NLU model in Persian language in shopping dialogue systems. They also offered a Name entity recognition model to realize product attributes. The list of products was crawled from DigiKala, the largest online shopping website in Persian. They also provide Persian conversations to train their models. However, there is no open dataset in the Persian language in the field of shopping. In this paper, we introduce PerSHOP, a dataset in which we gathered data on the shopping topic in Persian using crowd-sourcing. We develop this system using the MultiWOZ \citep{MultiWOZ} idea. We collected data using crowd-sourcing. Our work is in two parts. The first part is building a web-based application to gather conversations between customers and sellers. This system can automatically annotate some of the utterances in a rule-based approach. Then, we will complete annotating manually. So, it will be a benchmark for other researchers to train and evaluate their models on this dataset. The second part of our work is building some models using state-of-the-art methods with the help of language models, some tricks, and rules on top of this dataset and evaluating them in the natural language understanding task. This dataset contains 21925 utterances and 1061 dialogues. Each dialogue consists of a shopping basket, which often contains at least three products. These conversations were collected by 122 crowd workers for 30 days. Approximately, it took 200 hours for the annotating process. Table 1 shows a statistical comparison between this dataset and other relevant datasets in the field of task-oriented dialogue systems. As it is known, this dataset is not very large, but it is the first open dataset that is provided in the Persian language for shopping dialogue systems and can be used as a benchmark for future research. This dataset is available online for free.\footnote{https://github.com/keyvanmahmoudi/PerSHOP}
We built some Natural Language Understanding (NLU) models to provide a baseline for two different tasks: intent classification, and entity recognition in this dataset. We used some models like DIETClassifier \citep{DIETClassifier}, Conditional Random Field (CRF) model, and language models such as ParsBERT\citep{ParsBERT}, and laBSE\citep{laBSE}. Briefly, the contribution of this paper is as follows:

1) Development of a well-annotated Persian dataset in shopping dialogue systems. 

2) training some NLU models to two tasks: intent classification and entity recognition in this dataset. As far as we know, this is the first open Persian shopping dialogue dataset that is available online for other researchers.

\begin{table*}[htb]
\begin{tabular}{ p{3.7cm} | p{1cm} p{1.3cm} p{1cm} p{1.4cm} p{1.3cm} p{1.5cm} p{1.5cm} }
\hline
\textbf{Number of}	$\downarrow$ & \textbf{DSTC2} & \textbf{WOZ2.0} & \textbf{M2M}& \textbf{FRAMES}& \textbf{SGD}& \textbf{MultiWOZ} & \textbf{PerSHOP} \\
\hline					
Domains	& 1	& 1	& 2	& 3	& \textbf{16}	& 7	& 15 \\
Dialogues	& 1,612	& 600	& 1,500	& 1,369	& \textbf{16,142}	& 8,438	& 1,061 \\
Turns	& 23,354	& 4,472	& 14,796	& 19,986	& \textbf{329,964}	& 113,556	& 21,925 \\
 Mean turns per dialogue 	& 14.49	& 7.45	& 9.86	& 14.6	& 20.44	& 13.46	& \textbf{20.66} \\
Mean tokens per turn 	& 8.54	& 11.24	& 8.24	& 12.6	& 9.75	& \textbf{13.13}	& 8.9 \\
All tokens	& 199,431	& 50,264	& 121,977	& 251,867	& \textbf{3,217,149}	& 1,520,970	& 195,150 \\
Slots	& 8	& 4	& 13	& 61	& \textbf{214}	& 24	& 36 \\
Values	& 212	& 99	& 138	& 3,871	& \textbf{14,139}	& 4,510	& 7,939 \\
  \end{tabular}
\caption{Statistical comparison of this dataset with other datasets in the field of task-oriented dialogue systems}
  \label{tab:1}
\end{table*}
\section{Related works}
As we mentioned before, our work consists of two parts. The first part is about building a conversational dataset for shopping systems, and the second part is about introducing a benchmark model for NLU tasks. Regarding the first part, we should say that there are three approaches to building a conversational dataset. The first approach is human-to-human. It means that the user and the system are both humans, and they communicate with each other. This method is the most expensive but accurate way to collect conversations because both sides are humans, and their conversations are very near to real-world dialogues. One of the most famous datasets collected in this approach is the "Woz" dataset developed by \citep{WOZ}. First, in this method, an application to gather conversations between humans has been developed in the restaurant's domain through crowd-sourcing. Then, crowd-workers talk to each other via writing messages in this system. This system has two sides. One side is the end user who asks about restaurants in the Cambridge area of London, and the second side pretends to be the system and answers the user's asks according to a database that already has a list of the restaurants. Finally, these conversations are annotated by humans. Less than two years later, inspired by the same method, the MultiWOZ system was introduced by \citep{MultiWOZ}. This system implemented the same method as WOZ, this time for 7 different domains. It collected and annotated 10K dialogues through crowd-sourcing. These two datasets are both developed in the English language. \citep{CrossWOZ} created a dataset called CrossWOZ. This dataset is developed in the Chinese language and contains 6K dialogues. Another Chinese dataset named RiSAWOZ is proposed by \citep{risawoz}. This dataset is the largest Chinese dataset, and it is larger than MultiWOZ and contains more than 11K dialogues. \citep{X-RiSAWOZ} designed a mechanism in which, during four stages, it takes the dialogues in the RiSAWOZ dataset as input and translates them into the target languages. By this process, the RiSAWOZ dataset, which is in Chinese, has been translated into 5 languages: Hindi, English, Korean, French, and a mixed language of Hindi and English. Experiments show that this dataset which is called X-RiSAWOZ works very well in tasks such as dialogue state tracking. The second approach is human-to-machine. In this approach, humans interact with a dialogue system to collect dialogues.\citep{henderson-etal-2014-second} presented DSTC datasets with this approach. This dataset is appropriate for the dialogue state tracking task which is the brain of a task-oriented dialogue system. This category of methods is not as accurate as the first category, but it is considered a useful method for data collection. The third approach is machine-to-machine. In this category of methods, the end user and the chatbot are both machines. One of the famous datasets in this approach is presented by \citep{M2M}, and it is called M2M. In this method, at the first step, several domains are determined, and two machines talk to each other in these domains, and their conversations are saved. Then, in the second step, these conversations are annotated by humans manually. The most famous dataset created by this approach is the SGD dataset provided by \citep{SGD}. In this method, several services are first introduced, each of which has several intents and entities. Based on a predefined schema, the conversations will take place. Then, paraphrasing has been used to increase the volume of the dataset. This dataset is one of the largest datasets for task-oriented dialogue systems with 16K dialogues. \citep{casanueva-etal-2022-nlu} presented a Datatest for task-oriented dialogue systems in the domains of hotels and banking. This dataset has been collected and annotated by professional staff. It is called NLU++. They analyzed different models to understand natural language, and finally, they found a suitable model to perform NLU tasks for their dataset. \citep{MULTI3NLU++} expanded the work of NLU++ provided a new dataset called MULTI3NLU++. We know that the NLU++ dataset was presented in English. Using professional translators, they translated the conversations in this dataset into four other languages. These four languages are Spanish, Turkish, Amharic, and Marathi. In this way, they succeeded in creating a dataset for multilingual dialogue systems. Also, we can use this dataset for low-source languages. In general, their method can be useful for low-resource languages, because the datasets that exist in high-resource languages such as English can be translated into low-resource languages so high-quality models can be developed in these languages.

One of the problems in creating a dataset is that if we start from the zero point and collect datasets through methods such as crowd-sourcing, the amount of data obtained is not very large. As we know, one of the problems in the task-oriented dialogue systems is lack of data. We have already seen that \citep{MULTI3NLU++} have solved this problem by translating the dataset from a high-resource language to a low-resource language. However, the problem that remains here is that the cultural characteristics of languages and the people who use these languages to speak are different from each other. To solve this problem \citep{Cross-Lingual} created another multilingual dataset. In this dataset, they considered 4 different languages, and for each of these languages, according to their characteristics, they provide a plan that includes instructions for converting conversations from the source language to the target language. The multilingual dataset for Russian, Indonesian, Arabic, and Kiswahili languages was developed based on these guidelines. This dataset can be used for all dialogue system tasks.

In the scope of the second part of our work, there are some significant researches. In \citep{louvan-magnini-2020-recent} they deal with NLU tasks, intent classification, and slot filling which we will introduce our baseline model for these two tasks in this article too. They provide a survey for neural network-based models to address natural language understanding tasks in dialogue systems. \citep{gupta-etal-2019-casa} introduce an NLU model which is context-aware and it uses a self-attentive mechanism. It is called the CASA-NLU model. This model uses some information such as intents and slots that have been detected so far. They introduce an architecture with three components: text encoding, context fusion, and intent classification-slot filling component. They evaluated the model on some standard datasets and got better accuracy on natural language understanding tasks. \citep{xiao-etal-2021-end} proposed a method in the scope of shopping systems. This system is a combination of a search system and a dialogue system, and it uses the benefits of searching for a product in a dialogue system. This model has two transformers. One of them is for a traditional dialogue system, and another is for the products. they provide a profile for every product that exists in the product database. The combination of these two transformers helps the model to have a better understanding of the product that the customer wants. Considering there is not a lot of data to train a model in shopping dialogue systems, they also used a paraphrasing algorithm to deal with this problem. A constituency tree is used for utterance parsing in the domains with more conversations and transfers these utterances to domains with fewer conversations. \citep{LiveChat} designed an open domain dataset in Chinese called LiveChat. The reason we mention this dataset here is that the idea presented in it can also be used to build closed-domain datasets. They have done a 3-step mechanism to build this dataset. In the first stage, they collected videos of people's conversations. Then the audio in these videos has been extracted and converted to text by an ASR module. In the second stage, they have been made into utterances and answers. Finally, a model is built that takes an utterance from the input and provides a suitable answer for it.

\section{Dataset}
In this work, we build a dataset for shopping dialogue systems in the Persian language. This dataset can be used to train a seller dialogue agent.

\subsection{Dataset ontology}
At first, we built a database for a wide range of products. There are 750 products in 14 domains and 84 subdomains In this database. The list of these domains and subdomains of them is shown in Table 2. As shown in Table 2, there are many different products of a variety of categories in this database. Another domain named Chats exists in our dataset, that contains conversations such as Greetings, or Goodbye.
\begin{table*}[htb]
  \centering
\begin{tabular}{ |p{4.5cm}|p{10.5cm}|  }

\hline
\textbf{Domains}	& \textbf{Subdomains} \\
\hline
Beverages &	water/ non-alcoholic-beer/ juice/ coffee/ extracts/ nectar/ soda/ energy/ tea/ herbal \\
\hline
dairy	& cream/ cheese/ butter/ buttermilk/ milk/ yogurt/ curd/ ice-cream \\
\hline
Child	& baby-diaper/ baby-food/ baby-health/ baby-stuff/ baby-toy \\
\hline
Fruits and vegetables	& fruit/ summer/ vegetable \\
\hline
Dried, desserts, and sweets	& dessert/ palm/ candy/ dried/ cake \\
\hline
Grocery	& breakfast/ snacks/ additive/ oil/ noodles/ paste/ beans/ bread/ rice/ 
sweets/ powder/ cereal/ dough \\
\hline
Protein	& egg/ hen/ meat/ aquatic/ bologna \\
\hline
canned food	& conserve/ compote/ prepared-food/ freeze \\
\hline
Home and lifestyle	& kitchen/ disposable/ appliances/ bathroom/ hardware-tools/ gardening/ travel/ health-goods/ sport/ celebrate/ tobacco/ stationery \\
\hline
Vehicles	& car/ motorcycle/ bicycle \\
\hline
Fashion and clothing	& dress/ shoe/ accessory \\
\hline
Napkin and detergent	& napkin/ cleaning/ wash \\
\hline
Cosmetics	& makeup/ face/ hair/ body/ oral/ shave/ spray/ women-hygiene \\
\hline
Digital	& electricity/ electrical-appliance \\
\hline
Chats	& Greeting/ Goodbye/ Chitchat/ User-Confirm/ Deny/ Response-Deny/ Don’t-care/ Regret/ Response-Regret/ Basket/ Show-basket \\
\hline
  \end{tabular}
  \label{tab:2}
\caption{List of domains and subdomains}
\end{table*}

Figure 1 shows the number of utterances in each domain. As it is known, the number of conversations in the Chats domain is more than in the other domains. This is because dialogues usually start with greetings and end with goodbyes. Therefore, in most dialogues, this domain exists. The next domains in terms of the frequency of conversations are dairy products and beverages. One of the works that can be done in the next versions of this dataset is to do crowd-sourcing or use other methods to collect more conversations in domains that have fewer dialogues.

\begin{figure}[ht]
\centering
\includegraphics[width=0.5\textwidth]{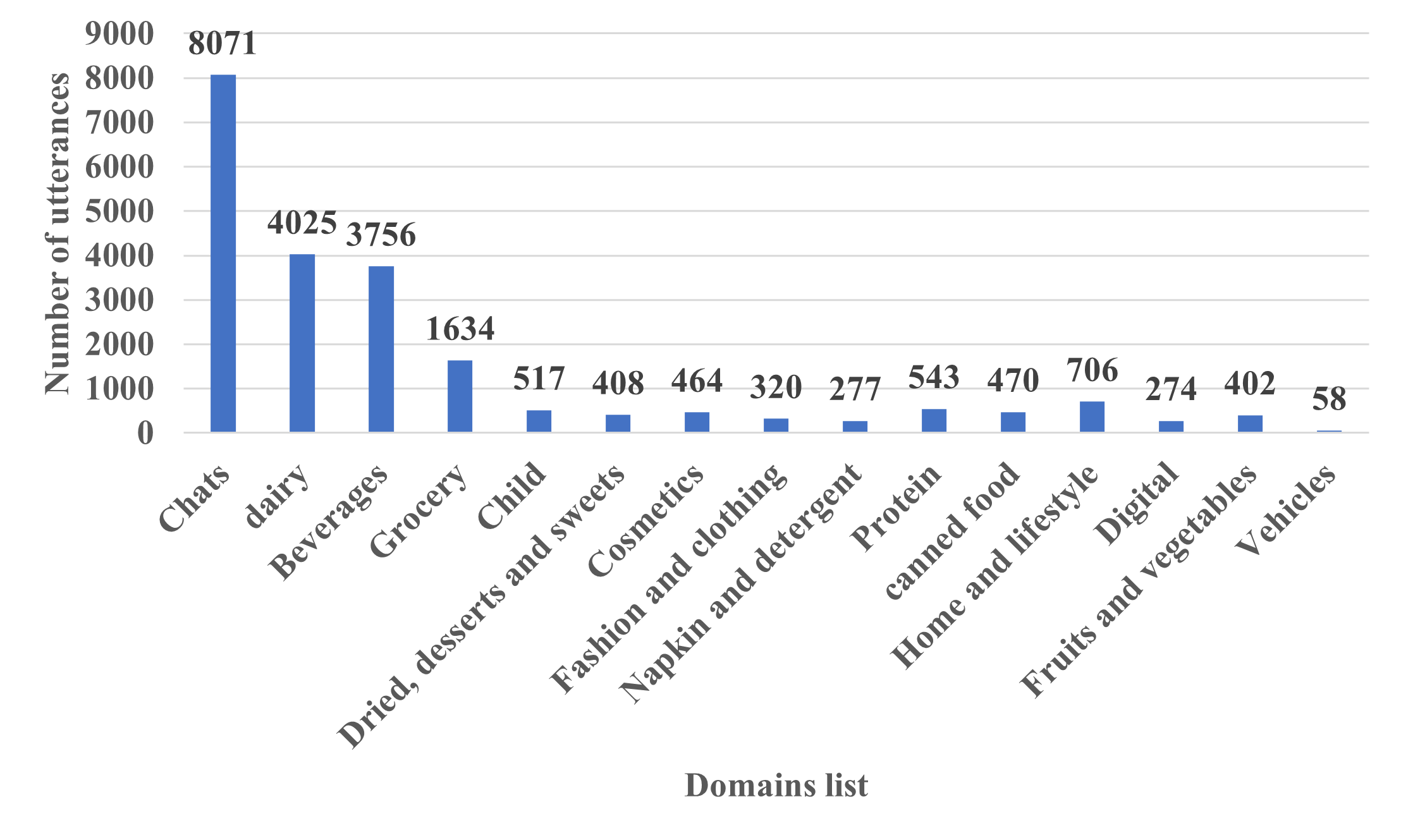}
\caption{Distribution of utterances in each domain}
\end{figure}

Also, there are 36 slots and each of them is a feature for different products. The list of these slots (features) is shown in Table 3. Some of these features are common among all products, such as the "brand" feature, which all products from any range and category have this feature. Some other features are common among many products, such as the "weight" feature, which many products have. On the other hand, some features are related to a small category of products, such as the "cheese-shape" feature, which is only related to cheese products, or the "energy-rank" feature, which is related to the category of electronic products such as refrigerators or washing machine. Table 3 provides a complete explanation of each of these slots, and in some cases, examples of the use of these slots are given.

\begin{table*}[ht!]
\centering
\begin{tabular}{ |p{2.5cm}|p{12.5cm}|}
\hline
\textbf{Slots (features)}	& \textbf{description} \\
\hline
product	& The general name of the product, such as milk. \\
\hline
super-product	& Domain or sub-domain of a product, such as dairy for milk. \\
\hline
category	& The general category of the product such as mutton in the meat product. \\
\hline
type	& The product type like honey type in a cream product. \\
\hline
fat	& Amount of the fat content of a food product such as milk or yogurt. \\
\hline
fresh	& Freshness or shelf life of an edible product such as milk. \\
\hline
size	& Descriptive size of a product, for example, large-size mineral water. \\
\hline
brand	& The brand of a product. \\
\hline
sub-brand	& Sub-brand of a famous brand. for example, Pompina is a sub-brand of the Kaleh brand. \\
\hline
volume	& The volume of a product such as milk. \\
\hline
cheese-shape	& Different shapes of cheeses. \\
\hline
weight	& The weight of a product. \\
\hline
flavor	& The taste of an edible product such as the taste of juice. \\
\hline
carbonate	& Carbonated or non-carbonated products such as soft drinks. \\
\hline
vegetables-of-buttermilk	& Buttermilk with vegetables or without vegetables. \\
\hline
family-drink	& Large size for beverage products such as soda or mineral water. \\
\hline
package	& The type of packaging of a product such as bottled soda. \\
\hline
bag	& The Brewable or urgency of products such as tea or coffee. \\
\hline
bag-number	& Number of bags for products such as tea bags. \\
\hline
size-number	& The numerical size of some products. Like a shoe in size 36. \\
\hline
color	& Descriptive color of a product. For example, a blue T-shirt. \\
\hline
color-number	& The numerical color of a product, for example, hair color. \\
\hline
country	& Country of manufacture of a product. \\
\hline
percent	& The percentage of something in a product, such as the percentage of meat in a hamburger. \\
\hline
height	& The height of a product. \\
\hline
width	& The width of a product. \\
\hline
length	& The length of a product. \\
\hline
diag	& The Diameter of a product of a product. \\
\hline
material	& Material used in a product for example silk in clothing. \\
\hline
pack-number	& The number in the packaging of products that are sold in bulk. For example, a pack involved 8 paper towels. \\
\hline
skin-type	& Types of skin for which the product is suitable, such as face cream. \\
\hline
hair-type	& Types of hair for which the product is suitable, such as shampoo. \\
\hline
gender	& The gender of the person who consumes the product, such as women's shoes. \\
\hline
energy-rank	& The energy consumption of an electrical product such as a refrigerator. \\
\hline
capacity	& The capacity of a product. for example, the capacity of a washing machine. \\
\hline
price	& The price of a product. \\
\hline
  \end{tabular}  
  \label{tab:3}
\caption{List of slots and the description of each of them}
\end{table*}

\subsection{Data collection method}
As mentioned in section 2, traditionally, a dataset can be created in 3 ways \citep{MultiWOZ}. The first method is that two artificial intelligence agents talk to each other, and Their conversations are stored. For example, in creating a shopping dataset, one of these agents is the seller, and the other one is the customer. The problem with this method is that the conversations obtained from this method are not similar to the real world, and two bots may not be able to produce the conversations we want. In the second method, one is human, and the other is a machine. In the shopping scenario, a human can be used as a customer, and an artificial intelligence agent can be used as a seller. In recent years, due to the development of Chat-GPT, it can be used for this purpose. This method is more accurate than the previous one, but the collected conversations are still different from real-world conversations. In the third method, both the speakers of the conversation are humans. Since two humans are talking to each other, this method is the most accurate way to build a dataset. The problem with this method is that is expensive and time-consuming. In this article, we used the third method because we intend to create an accurate and practical dataset that can be used as a baseline by other researchers and can be used in the real world too. We used crowd-sourcing to implement this method. Inspired by \citep{MultiWOZ}, we developed a web application. In the appendix, Figure 7 and Figure 8  show the environment of this application. It has two parts: The customer part and the seller part. In the customer part, crowd-workers can sign in and make purchases through a conversation. In the seller part, the seller can see the messages coming from the customer's side. the application automatically guides the seller to the page related to that product and shows the seller all the slots of this product and their values. it also has many default answers according to the user message which the seller can use or write the appropriate responses. 

\subsection{Data annotation}
To train a task-oriented dialogue system, we cannot give raw data to the model, so annotation is one of the important steps of our work. After collecting the data in the previous step through crowd-sourcing, we need to annotate them. We need to do two tasks for natural language understanding (NLU): intent classification and entity recognition. To train a model for each of these tasks we need annotated data. So, there are two kinds of annotating for conversations. the first is annotating each utterance as an intent.  For example, when the user says "I want low-fat milk" the user's intention is "purchase". The second is annotating the entities in this utterance.  It means the annotation of the product's features that are present in the user’s speech. For example, in the previous sentence, the value of "milk" will be annotated as "product" and the value of "low fat" will be annotated as "fat". For the intent classification task, we annotated data manually. In the entity recognition task, the application annotates a significant part of the data during the conversation in a rule-based manner. There is a code in our application to annotate conversations automatically.
 Figure 2 shows an example of the annotation done to specify the entities of a user's utterance. 
\begin{figure}[htb]
\centering
\includegraphics[width=0.5\textwidth]{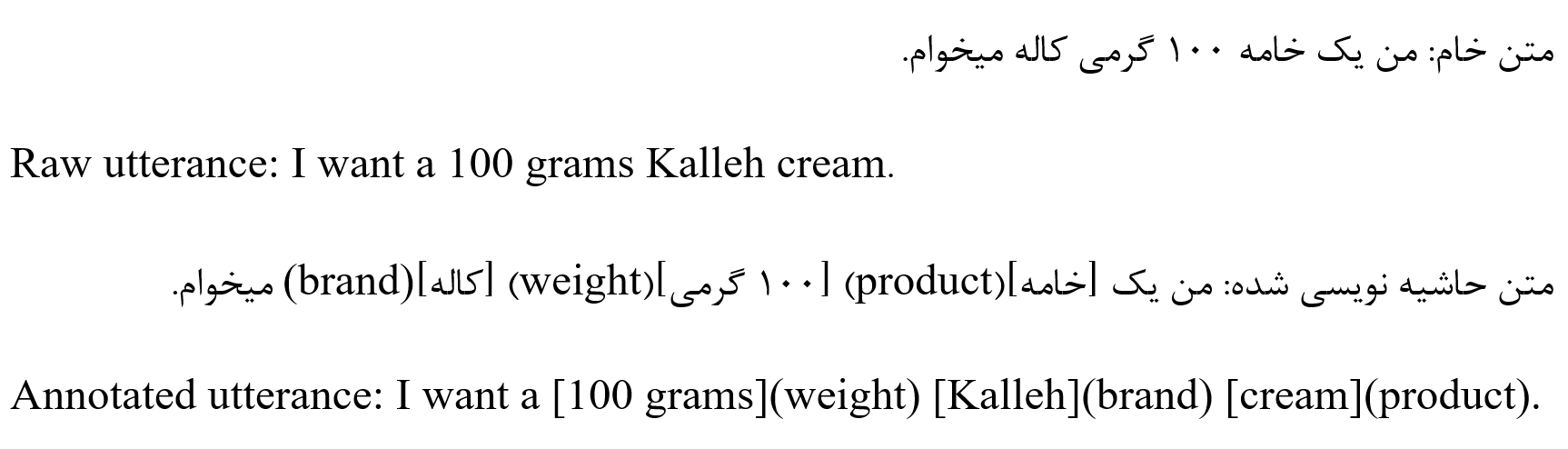}
\caption{An example of the annotation done in to specify the entities of a user's utterance.}
\end{figure}

In our application, when a user says an utterance, he/she has to select his/her domain, subdomain, and product from a drop-down list. So, after crowd-sourcing, we already have the product for each utterance. On the other side, we already have the slots and their values in our product database, so the code searches for the slot values in the product database and finds and annotates them in the user's utterance. All these processes do automatically in a second. Therefore, after crowd-sourcing and initial data collection, they are not completely raw data, and a significant part of them is automatically annotated by the application itself. Since this application outputs data in various formats such as spreadsheet format, we reviewed our dataset to find the probable wrong product selection by the users, wrong annotating by our automatic annotator system, or values which did not annotate by the system. We saw our system almost annotated more than half of the conversations automatically and we easily did the rest of the annotations manually. Finally, we achieved the fully annotated dataset.

\subsection{Dataset statistics}
Table 4 has two parts. The first part shows the number of customer utterances in each intent. The second part is the number of system utterances in each system action. To explain some of the intentions and actions in Table 4, we can consider a very simple scenario which is shown in Figure 3. When a customer wants to buy a product, his/her intent is called "Purchase". For example, if the customer utters the sentence "I want milk", in response to this sentence the system may ask about some of the product attributes, when the system asks "How much fat should it be?" The action of the system is called "Ask-slot". When the user says "High-fat is good" this is an "Inform" intent. If the system suggests a product according to what the customer has said so far, for example, if the system suggests a kind of milk, and asks from user if he/she wants to add it to his/her cart, the system action is called "Ask-product". If the customer confirms what the system suggested, for example, the customer says "Yes, I want this product", the customer's intention is "User-confirm", and in response to this sentence, the system will say something similar to this sentence: "The product with these features has been added to your basket. If you have anything else, let me know." This is an "Add-product" system action. On the other hand, if a customer says, for example, "No, I don't want this product", the customer's intention is "Deny", and the action of the system like "No problem. Can I do other services?" is called "Response-Deny". If the customer regrets buying the entire basket, the customer's intention is "Regret", For example when the user says "I easily gave up on buying this basket", the intention is regret. In this case, the system action is called "Response-Regret", for instance, the system says "No problem. Come anytime you want". 
\begin{figure}[ht]
\centering
\includegraphics[width=0.5\textwidth]{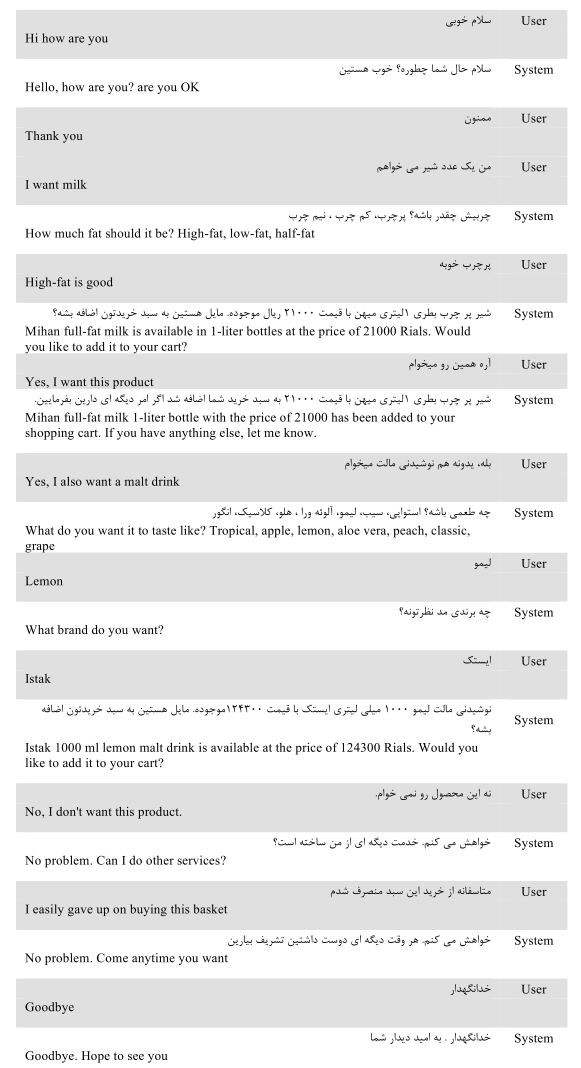}
\caption{A very simple scenario in this dataset}
\end{figure}
Some important statistics of this dataset are as follows. There are 11048 user utterances in 14 intents. Also, there are 10,877 utterances from the system (seller) in 9 different system actions. Overall, 21925 utterances were collected in 1061 dialogues. These conversations were collected by 122 crowd-workers for 30 days and each day for 4 hours. On average, each person has made at least 8.7 dialogues (shopping baskets). Approximately, it took 200 hours for the annotating process. Each dialog initially contains greetings. Then it has at least 3 product purchases and finally, it ends with goodbyes. Each dialogue has approximately 20 utterances between the customer and the system, with an average of 5 utterances for greetings and goodbyes and 15 utterances for the buying and selling process. According to the number of products that a customer buys in a shopping basket, there are 5 utterances to buy a product, half of them belong to the seller and the other half belong to the customer. In summary, it can be said that if a customer wants to buy a product, he/she should say 2 or 3 utterances about the desired product to add it to the basket. 

\begin{table}[t]
\begin{tabular}{ |p{4.5cm}|p{2.5cm}|  }
 \hline
\textbf{Customer intents}	& \textbf{ \# Utterances}  \\
 \hline
Purchase	& 3045	 \\
Inform	& 2389	  \\
User-Confirm	& 2776	 \\
Deny	& 199	  \\
Greeting	& 1411	 \\
Goodbye	& 1050	  \\
Chitchat	& 38	 \\
Regret	& 6	  \\
Basket	& 6	  \\
Prev-inform	& 15  \\
SuperProduct-ask	& 24 \\		
User-ask	& 76  \\	
Ask-detail	& 5  \\	
Don’t-care	& 8	 \\	
\hline
\textbf{Sum} & \textbf{11048}  \\
\hline
\hline
\textbf{System actions}	& \textbf{ \# Utterances}  \\
 \hline
Ask-slot &	 2616 \\
Ask-product	 & 2915 \\
Add-product	 & 2764 \\
Response-Deny	& 153 \\
Greeting	& 1336 \\
Goodbye	& 1061 \\
Chitchat	& 20 \\
Response-Regret	& 6 \\
Show-basket 	& 6 \\
\hline
\textbf{Sum} & \textbf{10877}\\
 \hline
 \hline
\textbf{Sum of all utterances} & \textbf{21925} \\
 \hline
\end{tabular}
\caption{Number of utterances of customer intents and system actions}
\end{table}

Figure 4 shows the number of turns in different dialogues. The distribution of turns in this dataset follows a normal distribution. As shown in Figure 4, conversations frequently have 18 or 20 turns. Figure 5 shows the number of tokens in each user and system utterances. Usually, the length of the system's statements is longer than that of the users. The reason is that users usually try to express their sentences quickly and concisely to reach their desire, which is to buy one or more specific products. On the other side, the system attempts to provide a suitable explanation for users and provide suggestions about product features to the customers so they can get the product they want more smoothly.
\begin{figure}[ht]
\centering
\includegraphics[width=0.5\textwidth]{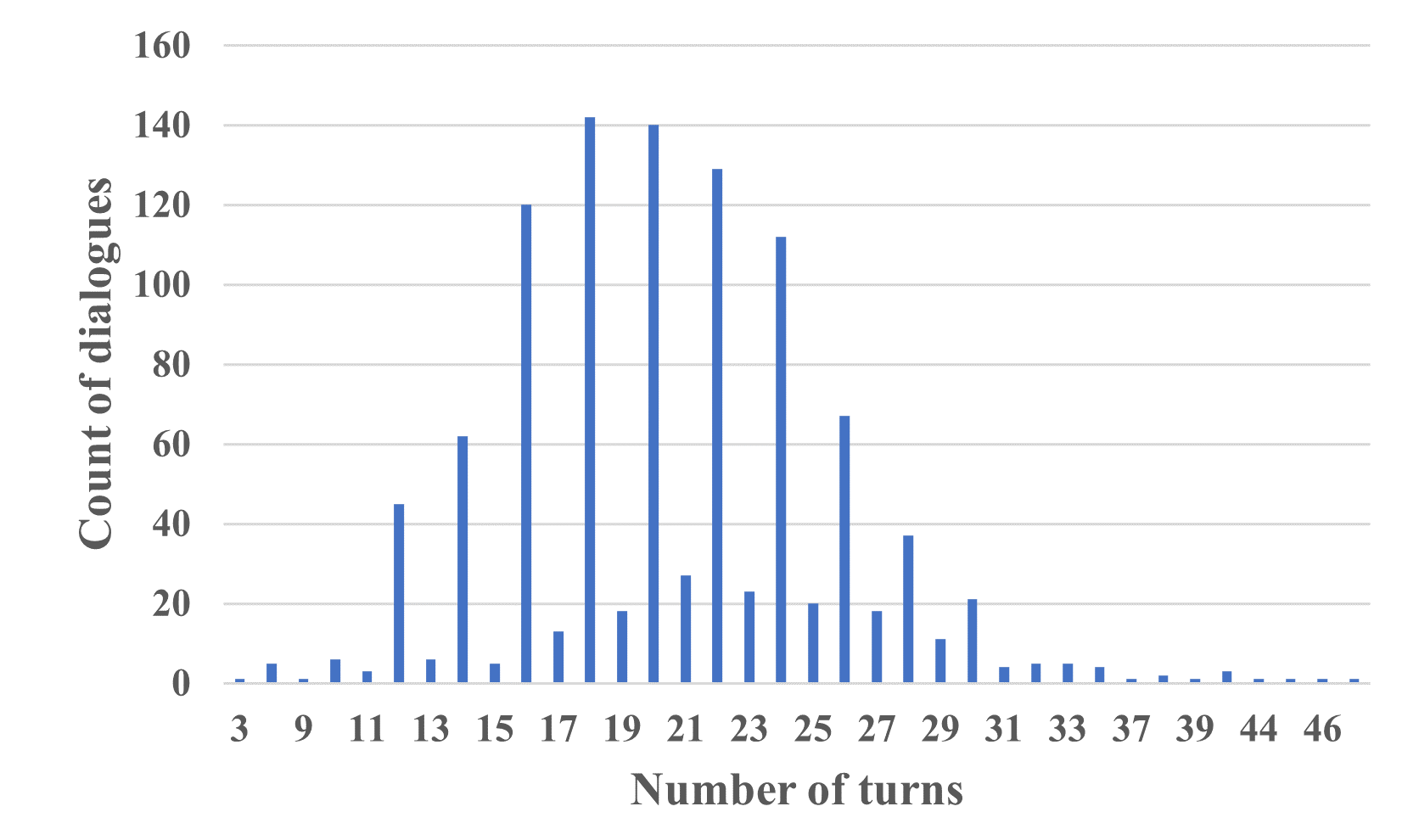}
\caption{Distribution of turns in dialogues}
\end{figure}

\begin{figure}[ht]
\centering
\includegraphics[width=0.5\textwidth]{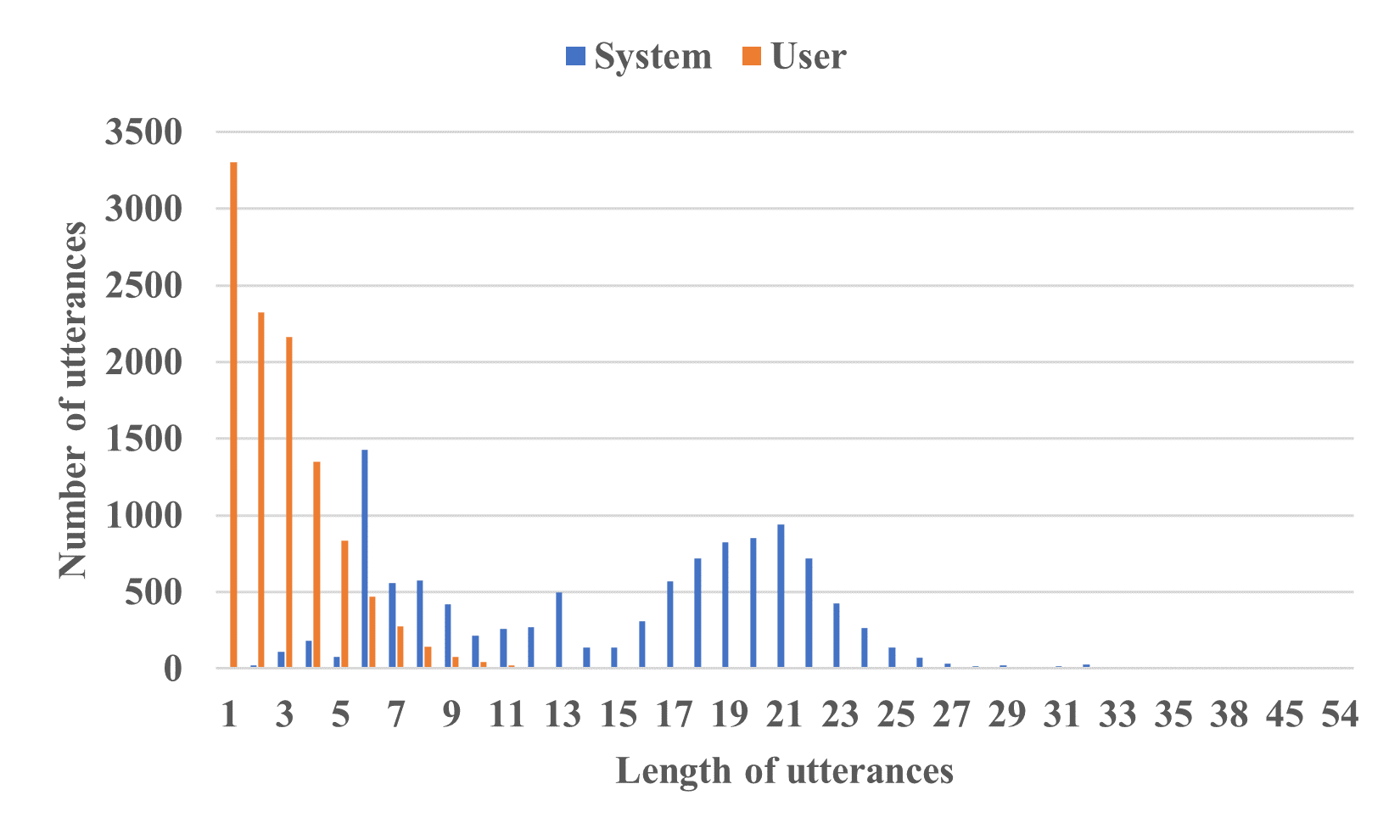}
\caption{Distribution of tokens in utterances}
\end{figure}

\section{Model \& Evaluation}
We want to build a Natural Language Understanding (NLU) model to provide a baseline for our dataset. We will build some NLU models for doing two different tasks: intent classification to understand what is the customer's intention in telling an utterance, and entity extraction to find out about the attributes of the product that the user wants. After, we will test and evaluate these models on our dataset.

\subsection{Natural Language Understanding (NLU) model}
First of all, we need to tokenize every single utterance. Considering the dataset's language is in Persian, we used Whitespace Tokenizer to do this. This tokenizer splits tokens based on a space between them.  Then, we need to convert each tokened utterance to feature vectors. To do this, we tested multiple language model weights as ParsBERT\citep{ParsBERT} Language model, and laBSE\citep{laBSE}. To classify user utterances to intents we used DIETClassifier \citep{DIETClassifier}. It stands for Dual Intent and Entity Transformer. We used this classifier because it does both of these tasks together and uses the same loss function for classification and sequence labeling. Given that in our work, classifying a text to intent and classifying each token of that text are naturally related, this classifier is a good choice. For example, assume the user states “Hello I want milk”. As we know, the intention of this text is “purchase”. If the classifier wrongly classifies the intention of this text as “Greetings”, but at the same time it detects “milk” as an attribute, it can find out the intention of this text is not “Greetings” because this intention has no milk. So, it can amend the intention of this sentence. We also used Fallback Classifier to handle situations when we were not sure about a classification. For example, when the intent classification value is less than a threshold, we use fallback to respond to the user with a default text. For example, “I am sorry. I did not understand you. Can you say that in another way?”. To extract entities from text, we used the Conditional Random Field (CRF) model addition to DIETClassifier. with these configurations, we built our models. Also, we used some tricks to increase the accuracy of the model. We used looking tables for obvious values to find them easily in the user text. For example, we put the list of a lot of brands in a looking table to find them in the user text. another thing that we do is we use a list of many synonyms of attributes and give that to the model. For example, the word "big" has synonyms like "large". Finally, our models have been built. 

\subsection{Evaluation}
We trained the DIETClassifier once with the ParsBERT language model and another time with the LaBSE language model and then evaluated it. We have used the same data for training and testing for both models. We have calculated the F1-Score for micro, macro, and weighted averages. Table 5 shows the results of these experiments. Both language models seem to have almost similar results and achieved a good score for intent classification.

\begin{table}[htb]
\begin{tabular}{ |p{3cm}|p{0.8cm}|p{0.9cm}|p{1.4cm}|}
\hline
 \textbf{Models} $\downarrow$ & \textbf{micro avg}	& \textbf{macro avg	} &\textbf{weighted avg} \\
 \hline
\scriptsize DIETClassifier + ParsBERT	& \textbf{91.14}	& 65.84	& \textbf{90.83} \\
\scriptsize DIETClassifier + LaBSE	& 90.79	& \textbf{76.08}	& 90.66 \\
 \hline
\end{tabular}
\caption{F1-Score of intent classification}
\end{table}

In Table 6, entity recognition is shown. In addition to the language models that we used for the intent classification, we used the Conditional Random Field (CRF) model. As can be seen, this model has a relatively better performance than the DIETClassifier for extracting entities.

\begin{table}[htb]
\begin{tabular}{ |p{3cm}|p{0.8cm}|p{0.9cm}|p{1.4cm}|}
 \hline
 \textbf{Models} $\downarrow$ & \textbf{micro avg}	& \textbf{macro avg	} &\textbf{weighted avg} \\
 \hline
 
\scriptsize DIETClassifier + ParsBERT	& 92.56	& \textbf{82.82}	& 92.4 \\

\scriptsize DIETClassifier + LaBSE	& 90.68	& 82.11	& 90.51 \\
\scriptsize CRF Entity Extractor	& \textbf{93.3}	& 79.98	& \textbf{92.9} \\
\hline
\end{tabular}
\caption{F1-Score of Entity Recognition}
\end{table}

\subsection{Error analysis}
There are some problems in the intention classification task. For example, when a customer says "full-fat milk." it is difficult to tell whether it is a "purchase" intention or an "inform" intention about the product. Also, there are issues with entity extraction. For example, when a customer says, "I want a mihan yogurt." if we don't have the brand "Mihan" in our train set, the model cannot recognize "Mihan" as a brand. To solve this problem, we use the lookup table for brands. Of course, we can also use lookup tables for other entities and increase the accuracy of entity extraction. Language models can help with entity extraction, too. For example, the language model can recognize the brand of a product. The language model may not have seen the brand name in the train set on which it was trained, but it has seen many sentences with the same structure. So, it might be to recognize the word "Mihan" as a brand in this sentence.

\section{Conclusions \& future works}
In this paper, we provide a dataset for shopping dialogue systems in the Persian language. We developed a web application to collect conversations between customers and the system through crowd-sourcing and automatically partially annotate them.  Then, we complete the annotation process manually. We also used some baseline models for NLU tasks which is intent classification and entity extraction. We performed experiments on our dataset and obtained good results as a benchmark for this dataset. For the future, considering that this dataset is still not very large, we suggest increasing the size of this dataset. For this, our method can be used, that is, using the crowd-source system. Another method is to use conversational paraphrasing for domains and intentions that have fewer dialogues, which can be done both automatically, like the SGD \citep{SGD} method, and manually. Another thing that can be done in this field is to translate the datasets in high-resource languages into low-resource languages such as Persian, so that we can increase the volume of the dataset significantly. This translation can be done manually by experts or automatically by some instructions.
For the task of building models, it is possible to make better models for natural language understanding using state-of-the-art methods. Also, considering that this dataset fully presents user and system conversations in an annotated form, a complete conversational model can be built on top of this dataset. A model that can completely replace humans and perform sales in practice. For this, pipeline models or end-to-end models can be used.

\nocite{*}
\section{References}\label{sec:reference}
\bibliographystyle{lrec-coling2024-natbib}
\bibliography{lrec-coling2024-example}

\begin{thebibliography}{0}
\expandafter\ifx\csname natexlab\endcsname\relax\def\natexlab#1{#1}\fi

\end{thebibliography}


\begin{thebibliography}{21}
\expandafter\ifx\csname natexlab\endcsname\relax\def\natexlab#1{#1}\fi

\bibitem[{Borhanifard et~al.(2020)Borhanifard, Basafa, Razavi, and Faili}]{Parsi}
Zeinab Borhanifard, Hossein Basafa, Seyedeh~Zahra Razavi, and Heshaam Faili. 2020.
\newblock \href {https://doi.org/10.1109/IKT51791.2020.9345639} {Persian language understanding in task-oriented dialogue system for online shopping}.
\newblock pages 79--84.

\bibitem[{Budzianowski et~al.(2018)Budzianowski, Wen, Tseng, Casanueva, Ultes, Ramadan, and Ga{\v{s}}i{\'c}}]{MultiWOZ}
Pawe{\l} Budzianowski, Tsung-Hsien Wen, Bo-Hsiang Tseng, I{\~n}igo Casanueva, Stefan Ultes, Osman Ramadan, and Milica Ga{\v{s}}i{\'c}. 2018.
\newblock \href {https://doi.org/10.18653/v1/D18-1547} {{M}ulti{WOZ} - a large-scale multi-domain {W}izard-of-{O}z dataset for task-oriented dialogue modelling}.
\newblock In \emph{Proceedings of the 2018 Conference on Empirical Methods in Natural Language Processing}, pages 5016--5026, Brussels, Belgium. Association for Computational Linguistics.

\bibitem[{Bunk et~al.(2020)Bunk, Varshneya, Vlasov, and Nichol}]{DIETClassifier}
Tanja Bunk, Daksh Varshneya, Vladimir Vlasov, and Alan Nichol. 2020.
\newblock \href {http://arxiv.org/abs/2004.09936} {Diet: Lightweight language understanding for dialogue systems}.

\bibitem[{Casanueva et~al.(2022)Casanueva, Vuli{\'c}, Spithourakis, and Budzianowski}]{casanueva-etal-2022-nlu}
Inigo Casanueva, Ivan Vuli{\'c}, Georgios Spithourakis, and Pawe{\l} Budzianowski. 2022.
\newblock \href {https://doi.org/10.18653/v1/2022.findings-naacl.154} {{NLU}++: A multi-label, slot-rich, generalisable dataset for natural language understanding in task-oriented dialogue}.
\newblock In \emph{Findings of the Association for Computational Linguistics: NAACL 2022}, pages 1998--2013, Seattle, United States. Association for Computational Linguistics.

\bibitem[{Conneau et~al.(2020)Conneau, Khandelwal, Goyal, Chaudhary, Wenzek, Guzm{\'a}n, Grave, Ott, Zettlemoyer, and Stoyanov}]{XLM-R}
Alexis Conneau, Kartikay Khandelwal, Naman Goyal, Vishrav Chaudhary, Guillaume Wenzek, Francisco Guzm{\'a}n, Edouard Grave, Myle Ott, Luke Zettlemoyer, and Veselin Stoyanov. 2020.
\newblock \href {https://doi.org/10.18653/v1/2020.acl-main.747} {Unsupervised cross-lingual representation learning at scale}.
\newblock In \emph{Proceedings of the 58th Annual Meeting of the Association for Computational Linguistics}, pages 8440--8451, Online. Association for Computational Linguistics.

\bibitem[{El~Asri et~al.(2017)El~Asri, Schulz, Sharma, Zumer, Harris, Fine, Mehrotra, and Suleman}]{Frames}
Layla El~Asri, Hannes Schulz, Shikhar Sharma, Jeremie Zumer, Justin Harris, Emery Fine, Rahul Mehrotra, and Kaheer Suleman. 2017.
\newblock \href {https://doi.org/10.18653/v1/W17-5526} {{F}rames: a corpus for adding memory to goal-oriented dialogue systems}.
\newblock In \emph{Proceedings of the 18th Annual {SIG}dial Meeting on Discourse and Dialogue}, pages 207--219, Saarbr{\"u}cken, Germany. Association for Computational Linguistics.

\bibitem[{Farahani et~al.(2021)Farahani, Gharachorloo, Farahani, and Manthouri}]{ParsBERT}
Mehrdad Farahani, Mohammad Gharachorloo, Marzieh Farahani, and Mohammad Manthouri. 2021.
\newblock \href {https://doi.org/10.1007/s11063-021-10528-4} {{ParsBERT}: Transformer-based model for persian language understanding}.
\newblock \emph{Neural Processing Letters}, 53(6):3831--3847.

\bibitem[{Feng et~al.(2022)Feng, Yang, Cer, Arivazhagan, and Wang}]{laBSE}
Fangxiaoyu Feng, Yinfei Yang, Daniel Cer, Naveen Arivazhagan, and Wei Wang. 2022.
\newblock \href {https://doi.org/10.18653/v1/2022.acl-long.62} {Language-agnostic {BERT} sentence embedding}.
\newblock In \emph{Proceedings of the 60th Annual Meeting of the Association for Computational Linguistics (Volume 1: Long Papers)}, pages 878--891, Dublin, Ireland. Association for Computational Linguistics.

\bibitem[{Gao et~al.(2023)Gao, Lian, Zhou, Fu, and Wang}]{LiveChat}
Jingsheng Gao, Yixin Lian, Ziyi Zhou, Yuzhuo Fu, and Baoyuan Wang. 2023.
\newblock \href {https://doi.org/10.18653/v1/2023.acl-long.858} {{L}ive{C}hat: A large-scale personalized dialogue dataset automatically constructed from live streaming}.
\newblock In \emph{Proceedings of the 61st Annual Meeting of the Association for Computational Linguistics (Volume 1: Long Papers)}, pages 15387--15405, Toronto, Canada. Association for Computational Linguistics.

\bibitem[{Gupta et~al.(2019)Gupta, Zhang, Lalwani, and Diab}]{gupta-etal-2019-casa}
Arshit Gupta, Peng Zhang, Garima Lalwani, and Mona Diab. 2019.
\newblock \href {https://doi.org/10.18653/v1/D19-1127} {{CASA}-{NLU}: Context-aware self-attentive natural language understanding for task-oriented chatbots}.
\newblock In \emph{Proceedings of the 2019 Conference on Empirical Methods in Natural Language Processing and the 9th International Joint Conference on Natural Language Processing (EMNLP-IJCNLP)}, pages 1285--1290, Hong Kong, China. Association for Computational Linguistics.

\bibitem[{Henderson et~al.(2014)Henderson, Thomson, and Williams}]{henderson-etal-2014-second}
Matthew Henderson, Blaise Thomson, and Jason~D. Williams. 2014.
\newblock \href {https://doi.org/10.3115/v1/W14-4337} {The second dialog state tracking challenge}.
\newblock In \emph{Proceedings of the 15th Annual Meeting of the Special Interest Group on Discourse and Dialogue ({SIGDIAL})}, pages 263--272, Philadelphia, PA, U.S.A. Association for Computational Linguistics.

\bibitem[{Louvan and Magnini(2020)}]{louvan-magnini-2020-recent}
Samuel Louvan and Bernardo Magnini. 2020.
\newblock \href {https://doi.org/10.18653/v1/2020.coling-main.42} {Recent neural methods on slot filling and intent classification for task-oriented dialogue systems: A survey}.
\newblock In \emph{Proceedings of the 28th International Conference on Computational Linguistics}, pages 480--496, Barcelona, Spain (Online). International Committee on Computational Linguistics.

\bibitem[{Majewska et~al.(2023)Majewska, Razumovskaia, Ponti, Vuli{\'c}, and Korhonen}]{Cross-Lingual}
Olga Majewska, Evgeniia Razumovskaia, Edoardo~M. Ponti, Ivan Vuli{\'c}, and Anna Korhonen. 2023.
\newblock \href {https://doi.org/10.1162/tacl_a_00539} {Cross-lingual dialogue dataset creation via outline-based generation}.
\newblock \emph{Transactions of the Association for Computational Linguistics}, 11:139--156.

\bibitem[{Moghe et~al.(2023)Moghe, Razumovskaia, Guillou, Vuli{\'c}, Korhonen, and Birch}]{MULTI3NLU++}
Nikita Moghe, Evgeniia Razumovskaia, Liane Guillou, Ivan Vuli{\'c}, Anna Korhonen, and Alexandra Birch. 2023.
\newblock \href {https://doi.org/10.18653/v1/2023.findings-acl.230} {{M}ulti3{NLU}++: A multilingual, multi-intent, multi-domain dataset for natural language understanding in task-oriented dialogue}.
\newblock In \emph{Findings of the Association for Computational Linguistics: ACL 2023}, pages 3732--3755, Toronto, Canada. Association for Computational Linguistics.

\bibitem[{Moradshahi et~al.(2023)Moradshahi, Shen, Bali, Choudhury, de~Chalendar, Goel, Kim, Kodali, Kumaraguru, Semmar, Semnani, Seo, Seshadri, Shrivastava, Sun, Yadavalli, You, Xiong, and Lam}]{X-RiSAWOZ}
Mehrad Moradshahi, Tianhao Shen, Kalika Bali, Monojit Choudhury, Gael de~Chalendar, Anmol Goel, Sungkyun Kim, Prashant Kodali, Ponnurangam Kumaraguru, Nasredine Semmar, Sina Semnani, Jiwon Seo, Vivek Seshadri, Manish Shrivastava, Michael Sun, Aditya Yadavalli, Chaobin You, Deyi Xiong, and Monica Lam. 2023.
\newblock \href {https://doi.org/10.18653/v1/2023.findings-acl.174} {{X}-{R}i{SAWOZ}: High-quality end-to-end multilingual dialogue datasets and few-shot agents}.
\newblock In \emph{Findings of the Association for Computational Linguistics: ACL 2023}, pages 2773--2794, Toronto, Canada. Association for Computational Linguistics.

\bibitem[{Quan et~al.(2020)Quan, Zhang, Cao, Li, and Xiong}]{risawoz}
Jun Quan, Shian Zhang, Qian Cao, Zizhong Li, and Deyi Xiong. 2020.
\newblock \href {https://doi.org/10.18653/v1/2020.emnlp-main.67} {{R}i{SAWOZ}: A large-scale multi-domain {W}izard-of-{O}z dataset with rich semantic annotations for task-oriented dialogue modeling}.
\newblock In \emph{Proceedings of the 2020 Conference on Empirical Methods in Natural Language Processing (EMNLP)}, pages 930--940, Online. Association for Computational Linguistics.

\bibitem[{Rastogi et~al.(2019)Rastogi, Zang, Sunkara, Gupta, and Khaitan}]{SGD}
Abhinav Rastogi, Xiaoxue Zang, Srinivas Sunkara, Raghav Gupta, and Pranav Khaitan. 2019.
\newblock \href {http://arxiv.org/abs/1909.05855} {Towards scalable multi-domain conversational agents: The schema-guided dialogue dataset}.
\newblock \emph{CoRR}, abs/1909.05855.

\bibitem[{Shah et~al.(2018)Shah, Hakkani-T{\"u}r, T{\"u}r, Rastogi, Bapna, Kennard, and Heck}]{M2M}
Pararth Shah, Dilek~Z. Hakkani-T{\"u}r, G{\"o}khan T{\"u}r, Abhinav Rastogi, Ankur Bapna, Neha~Nayak Kennard, and Larry Heck. 2018.
\newblock \href {https://api.semanticscholar.org/CorpusID:24831380} {Building a conversational agent overnight with dialogue self-play}.
\newblock \emph{ArXiv}, abs/1801.04871.

\bibitem[{Wen et~al.(2017)Wen, Vandyke, Mrk{\v{s}}i{\'c}, Ga{\v{s}}i{\'c}, Rojas-Barahona, Su, Ultes, and Young}]{WOZ}
Tsung-Hsien Wen, David Vandyke, Nikola Mrk{\v{s}}i{\'c}, Milica Ga{\v{s}}i{\'c}, Lina~M. Rojas-Barahona, Pei-Hao Su, Stefan Ultes, and Steve Young. 2017.
\newblock \href {https://aclanthology.org/E17-1042} {A network-based end-to-end trainable task-oriented dialogue system}.
\newblock In \emph{Proceedings of the 15th Conference of the {E}uropean Chapter of the Association for Computational Linguistics: Volume 1, Long Papers}, pages 438--449, Valencia, Spain. Association for Computational Linguistics.

\bibitem[{Xiao et~al.(2021)Xiao, Ma, Dong, Mart{\'\i}nez-G{\'o}mez, Zalmout, Chen, Zhao, He, and Jin}]{xiao-etal-2021-end}
Liqiang Xiao, Jun Ma, Xin~Luna Dong, Pascual Mart{\'\i}nez-G{\'o}mez, Nasser Zalmout, Wei Chen, Tong Zhao, Hao He, and Yaohui Jin. 2021.
\newblock \href {https://doi.org/10.18653/v1/2021.emnlp-main.280} {End-to-end conversational search for online shopping with utterance transfer}.
\newblock In \emph{Proceedings of the 2021 Conference on Empirical Methods in Natural Language Processing}, pages 3477--3486, Online and Punta Cana, Dominican Republic. Association for Computational Linguistics.

\bibitem[{Zhu et~al.(2020)Zhu, Huang, Zhang, Zhu, and Huang}]{CrossWOZ}
Qi~Zhu, Kaili Huang, Zheng Zhang, Xiaoyan Zhu, and Minlie Huang. 2020.
\newblock \href {https://doi.org/10.1162/tacl_a_00314} {Crosswoz: A large-scale chinese cross-domain task-oriented dialogue dataset}.
\newblock \emph{Transactions of the Association for Computational Linguistics}, 8:281--295.

\end{thebibliography}
\bibliographystylelanguageresource{lrec-coling2024-natbib}
\bibliographylanguageresource{languageresource}

\appendix
{\clearpage}
\section{Appendix}
\begin{figure}[htb]
\centering
\includegraphics[width=1\textwidth]{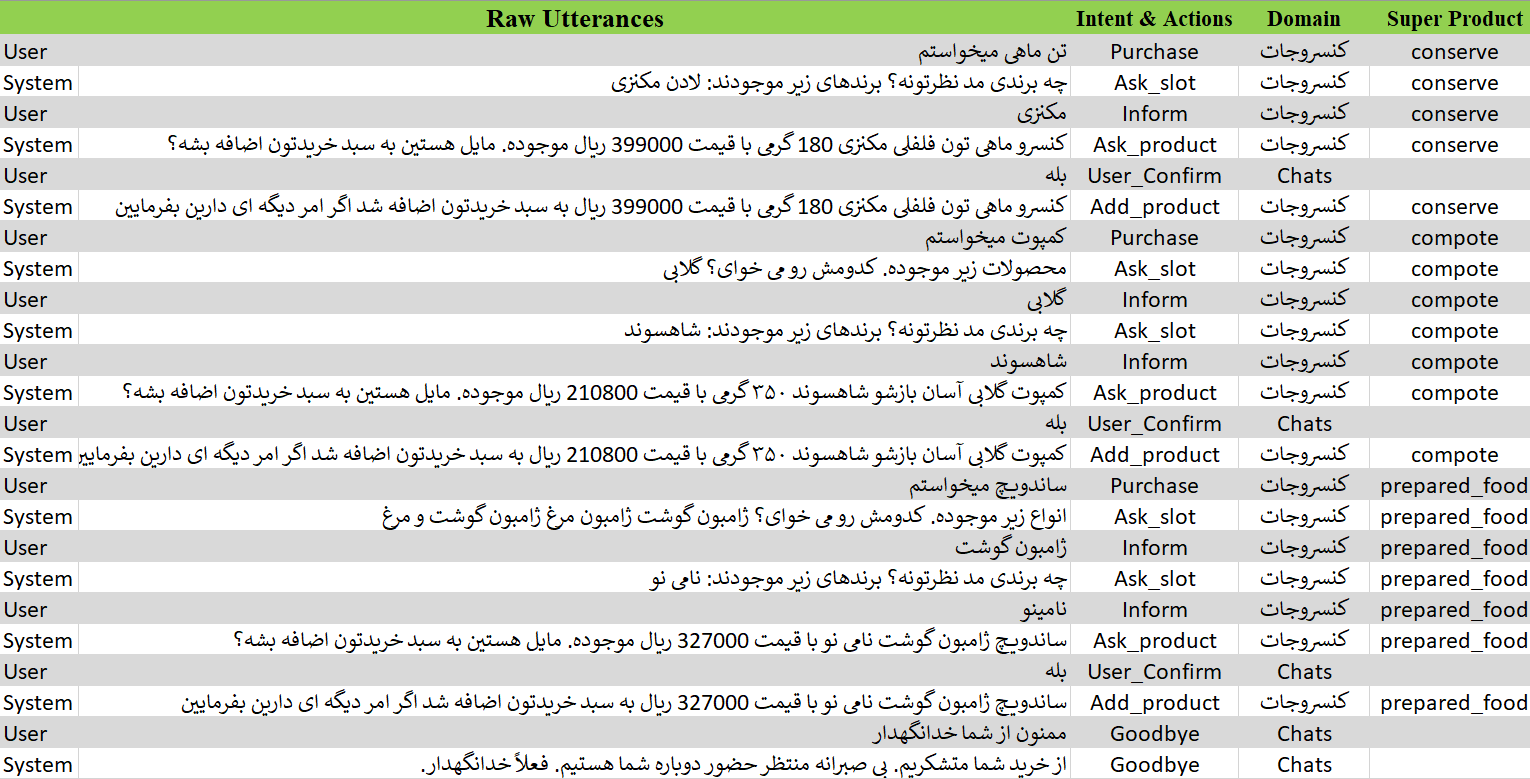}
\captionsetup{width=11cm}
\caption{An example of a dialogue between a user and the system}
\end{figure}

\begin{figure}[htb]
\centering
\includegraphics[width=1\textwidth]{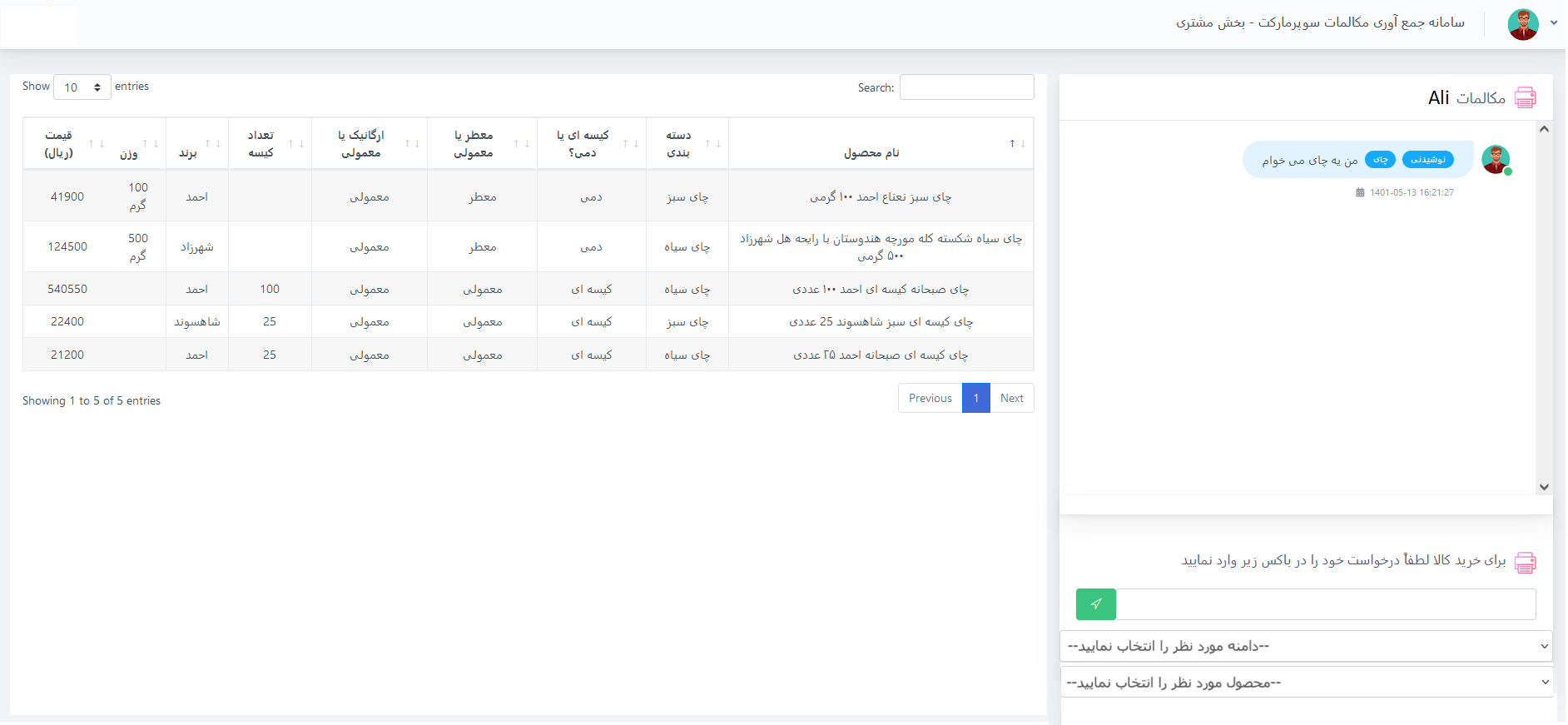}
\captionsetup{width=11cm}
\caption{A view from the user side of our crowd-sourcing web application}
\end{figure}

\clearpage
\begin{figure}[htb]
\centering
\includegraphics[width=1\textwidth]{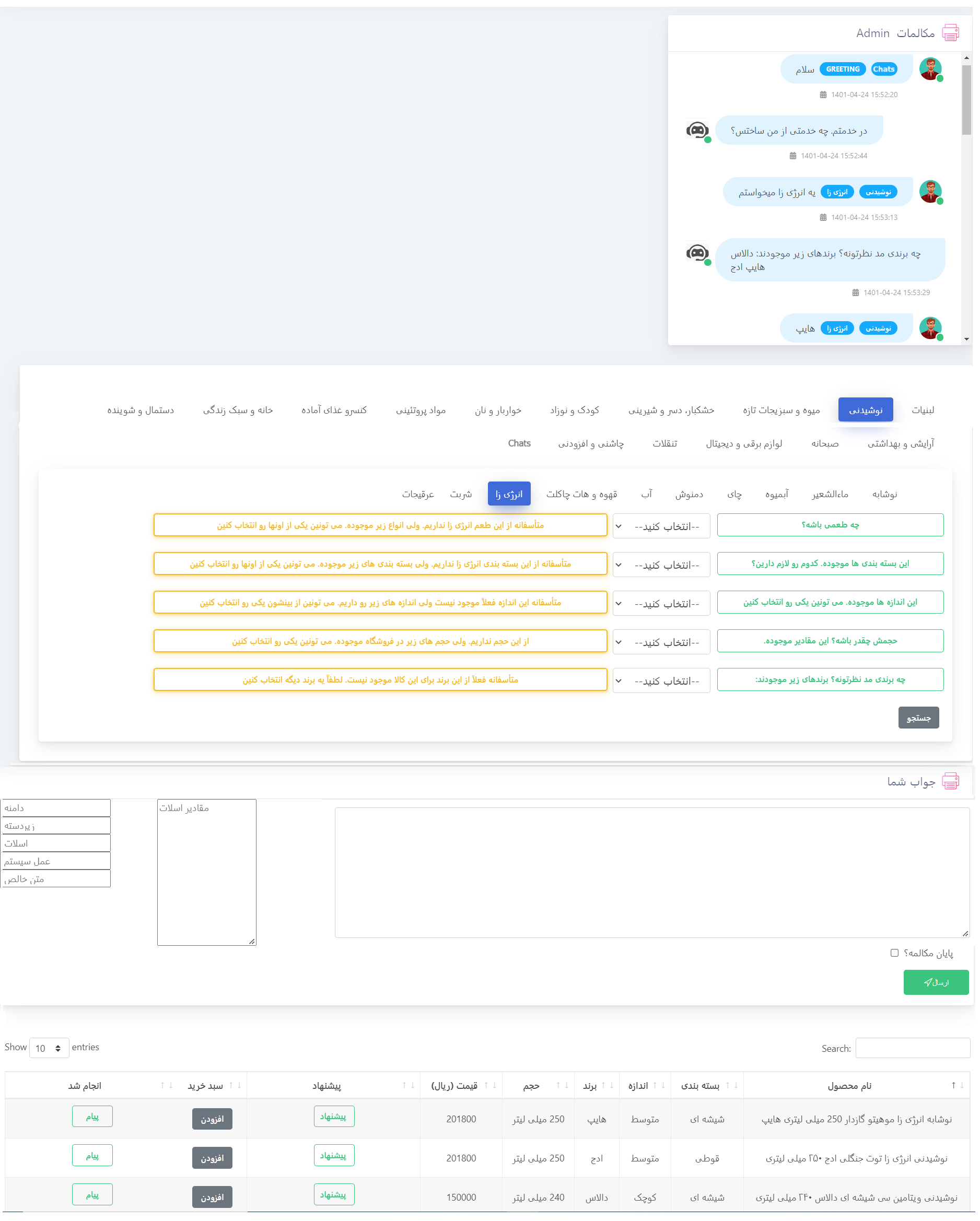}
\captionsetup{width=11.5cm}
\caption{A view from the system side of our crowd-sourcing web application}
\end{figure}

\end{document}